\documentclass[letterpaper]{article} 
\usepackage{aaai2027}
\usepackage[hyphens]{url}  
\usepackage{graphicx} 
\usepackage{natbib}  
\usepackage{caption} 
\usepackage{algorithm}
\usepackage{amssymb}

\usepackage{newfloat}
\usepackage{listings}
\DeclareCaptionStyle{ruled}{labelfont=normalfont,labelsep=colon,strut=off} 
\floatstyle{ruled}
\newfloat{listing}{tb}{lst}{}
\floatname{listing}{Listing}

\usepackage{booktabs}
\usepackage{amsmath}
\usepackage{multirow}
\usepackage{pifont}
\usepackage{algpseudocode}

\title{VFAD: Variational Semantic Prompting Meets Frequency-Adaptive Representation Learning for Zero-Shot Anomaly Detection}

\author{
    Peng Chen,
    Kaige Li,
    Wei Wang,
    Mingbo Yang,
    Wenqiang Wang,
    Li Shen,\\
    Fangjun Huang,
    Chao Huang\thanks{Corresponding author. Email: huangch253@mail.sysu.edu.cn}
}

\affiliations{
    School of Cyber Science and Technology, Shenzhen Campus of Sun Yat-sen University, Shenzhen, China
}

\begin{document}

\maketitle

\begin{abstract}
Zero-shot anomaly detection (ZSAD) aims to detect and localize anomalies in unseen categories without access to target-specific training data. Although recent CLIP-based methods have demonstrated promising generalization through vision-language alignment, they remain limited in capturing diverse anomaly semantics and subtle local variations. To address these limitations, we propose VFAD, a unified framework that combines variational semantic prompting with frequency-adaptive representation learning. Specifically, we introduce a Variational Semantic Prompt Extractor (VSPE), which adaptively aggregates anomaly-relevant local semantics from dense patch tokens and regularizes them through a variational information bottleneck, thereby incorporating fine-grained visual cues and enabling more precise cross-modal alignment. Furthermore, we develop a Frequency-Adaptive Representation Aggregation (FARA) module that leverages wavelet-based frequency decomposition and frequency-specific expert aggregation to enhance anomaly-discriminative visual representations. By jointly strengthening semantic guidance and visual representation learning, VFAD improves both anomaly discrimination and fine-grained localization. Extensive experiments on 13 industrial and medical benchmarks demonstrate that VFAD consistently outperforms existing state-of-the-art ZSAD methods across diverse anomaly scenarios. The code will be publicly available upon publication.
\end{abstract}


\section{Introduction}

Anomaly detection aims to identify samples or spatial regions that deviate from normal patterns, with broad applications in industrial inspection and medical image analysis~\cite{gao2026adaptclip,yao2026mmr}. Conventional approaches are typically developed under category-specific settings, requiring target-category data and limiting their generalization to unseen categories~\cite{xu2025towards,hou2026visualad}. Since anomalous samples are rare, diverse, and expensive to annotate, collecting sufficient data for emerging categories remains challenging~\cite{lv2025one,wang2026mau}. Zero-shot anomaly detection (ZSAD) addresses these challenges by transferring anomaly knowledge learned from auxiliary data to unseen categories without requiring any target-category training samples.
\begin{figure}[t]
    \centering
    \includegraphics[width=\linewidth]{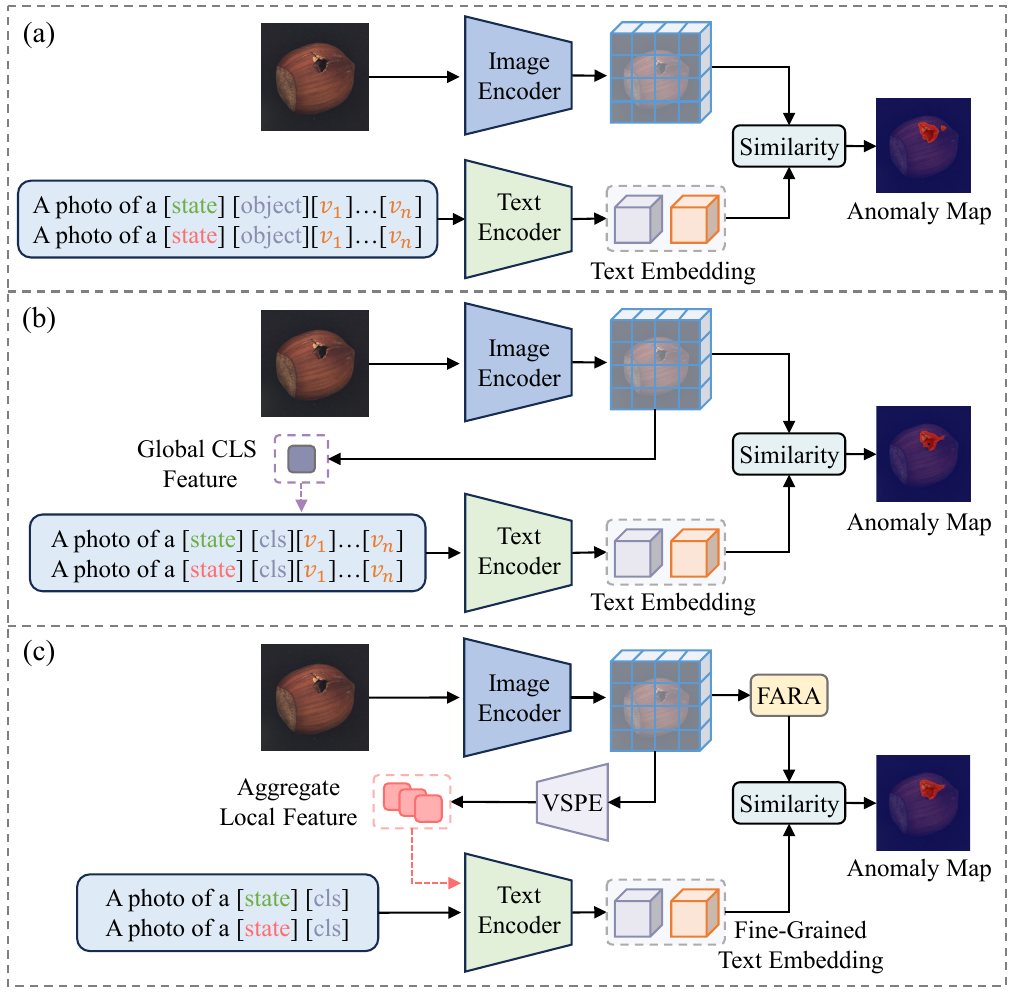}
    \caption{Comparison of existing CLIP-based ZSAD methods and our VFAD. (a) Rely on general text prompts. (b) Adapt global visual features for prompt enhancement. (c) VFAD extracts anomaly-relevant local visual semantics as dynamic prompts and leverages frequency-adaptive representations for more discriminative cross-modal alignment.}
    \label{fig:motivation}
\end{figure}

The emergence of large-scale vision-language models (VLMs), particularly CLIP~\cite{radford2021learning}, has significantly advanced ZSAD by providing transferable visual-semantic representations learned from massive image-text pretraining~\cite{qu2025dictas,jin2026reasoning}. Early approaches, such as WinCLIP~\cite{jeong2023winclip}, leverage handcrafted normal and abnormal prompts together with multi-scale visual features for anomaly localization. Subsequent studies improve CLIP-based ZSAD from complementary perspectives, including semantic prompt learning and visual feature adaptation. AnomalyCLIP~\cite{zhou2024anomalyclip} introduces object-agnostic learnable prompts to capture category-independent anomaly semantics, while Bayes-PFL~\cite{qu2025bayesian} enhances prompt learning through probabilistic prompt modeling. More recently, MoECLIP~\cite{park2026moeclip} explores patch-level expert adaptation to obtain more specialized visual representations. Despite these advances, existing CLIP-based ZSAD methods still face challenges in achieving robust semantic understanding and fine-grained anomaly localization.

These limitations mainly arise from two aspects. First, existing methods struggle to capture diverse and ambiguous anomaly semantics. As illustrated in Figure~\ref{fig:motivation}(a), most approaches rely on handcrafted templates or deterministic learnable prompts, which represent normal and anomalous states with fixed semantic embeddings. However, anomalies exhibit substantial variations across categories, making such prompts insufficient to capture their semantic diversity and uncertainty. Second, existing methods remain limited in modeling fine-grained visual variations. Since CLIP is pretrained for global image-text alignment, it tends to emphasize dominant object-level semantics while overlooking subtle anomaly cues. As shown in Figure~\ref{fig:motivation}(b), global visual adaptation may introduce irrelevant object and background information, while uniform patch transformation ignores the semantic heterogeneity of different regions and the complementary roles of structural and textural cues.

To address these challenges, we propose VFAD, a novel framework that integrates Variational Semantic Prompting and Frequency-Adaptive Representation Learning for zero-shot anomaly detection. As shown in Figure~\ref{fig:motivation}(c), our key idea is to enhance ZSAD by jointly improving anomaly-aware semantic guidance and fine-grained visual representation learning. Specifically, we introduce a Variational Semantic Prompt Extractor (VSPE), which dynamically extracts anomaly-relevant local semantics from dense patch tokens and regularizes them through a variational information bottleneck, capturing fine-grained semantic context while reducing reliance on global visual representations. Furthermore, we develop a Frequency-Adaptive Representation Aggregation (FARA) module that decomposes visual features into complementary low- and high-frequency components and adaptively aggregates them through frequency-specific experts, enhancing texture-sensitive representations while preserving structural information for accurate anomaly localization. Our contributions are summarized as follows:

\begin{itemize}
    \item We propose VFAD, a novel zero-shot anomaly detection framework that jointly explores variational semantic prompting and frequency-adaptive representation learning for enhanced fine-grained anomaly understanding.
    
    \item We introduce VSPE, which extracts anomaly-relevant local semantics from dense patch tokens and injects them into prompt learning under variational regularization.
    
    \item We develop FARA, a frequency-adaptive representation aggregation module that leverages wavelet-based frequency decomposition and expert aggregation to enhance anomaly-discriminative visual representations.
    
    \item Extensive experiments on 13 industrial and medical benchmarks demonstrate that VFAD achieves superior zero-shot generalization across unseen categories.
\end{itemize}

\section{Related Work}

\subsection{Zero-Shot Anomaly Detection}

Zero-shot anomaly detection (ZSAD) aims to identify abnormal samples or localize anomalous regions in unseen categories without requiring target-category training samples~\cite{chen2026dyc}. Recent CLIP-based approaches have significantly advanced ZSAD by leveraging vision-language alignment for transferable anomaly detection and localization~\cite{zhang2026dlvp}. WinCLIP~\cite{jeong2023winclip} pioneers the application of CLIP to ZSAD by matching handcrafted normal and abnormal prompts with multi-scale visual features for anomaly localization. AnomalyCLIP~\cite{zhou2024anomalyclip} further introduces object-agnostic learnable prompts to capture category-independent anomaly semantics. Subsequent methods further enhance CLIP-based ZSAD from different perspectives, including visual-context conditioning in AdaCLIP~\cite{cao2024adaclip}, probabilistic prompt modeling in Bayes-PFL~\cite{qu2025bayesian}, discriminative text anchors in AA-CLIP~\cite{ma2025aa}, and patch-specialized expert adaptation in MoECLIP~\cite{park2026moeclip}. Despite these advances, existing approaches still face challenges in modeling diverse anomaly semantics and learning fine-grained visual representations.

\subsection{Prompt Learning}

Prompt learning has emerged as an effective strategy for adapting large-scale vision-language models to downstream tasks~\cite{shao2026promptmoe}. CoOp~\cite{zhou2022learning} replaces handcrafted templates with learnable context tokens, CoCoOp~\cite{zhou2022conditional} conditions prompts on visual features to improve generalization, and GalLoP~\cite{lafon2024gallop} jointly learns global and local prompts to incorporate fine-grained visual cues. In ZSAD, VCP-CLIP~\cite{qu2024vcp} introduces visual context for image-conditioned prompting, while FE-CLIP~\cite{gong2025fe} complements prompt adaptation with frequency-aware visual representations. However, existing prompt learning strategies often rely on deterministic optimization and remain limited in capturing diverse and ambiguous anomaly semantics. In contrast, our Variational Semantic Prompt Extractor distills anomaly-relevant semantics from dense patch tokens through a variational information bottleneck, enabling more discriminative fine-grained cross-modal alignment.

\section{Methodology}

\begin{figure*}[t]
    \centering
    \includegraphics[width=0.96\linewidth]{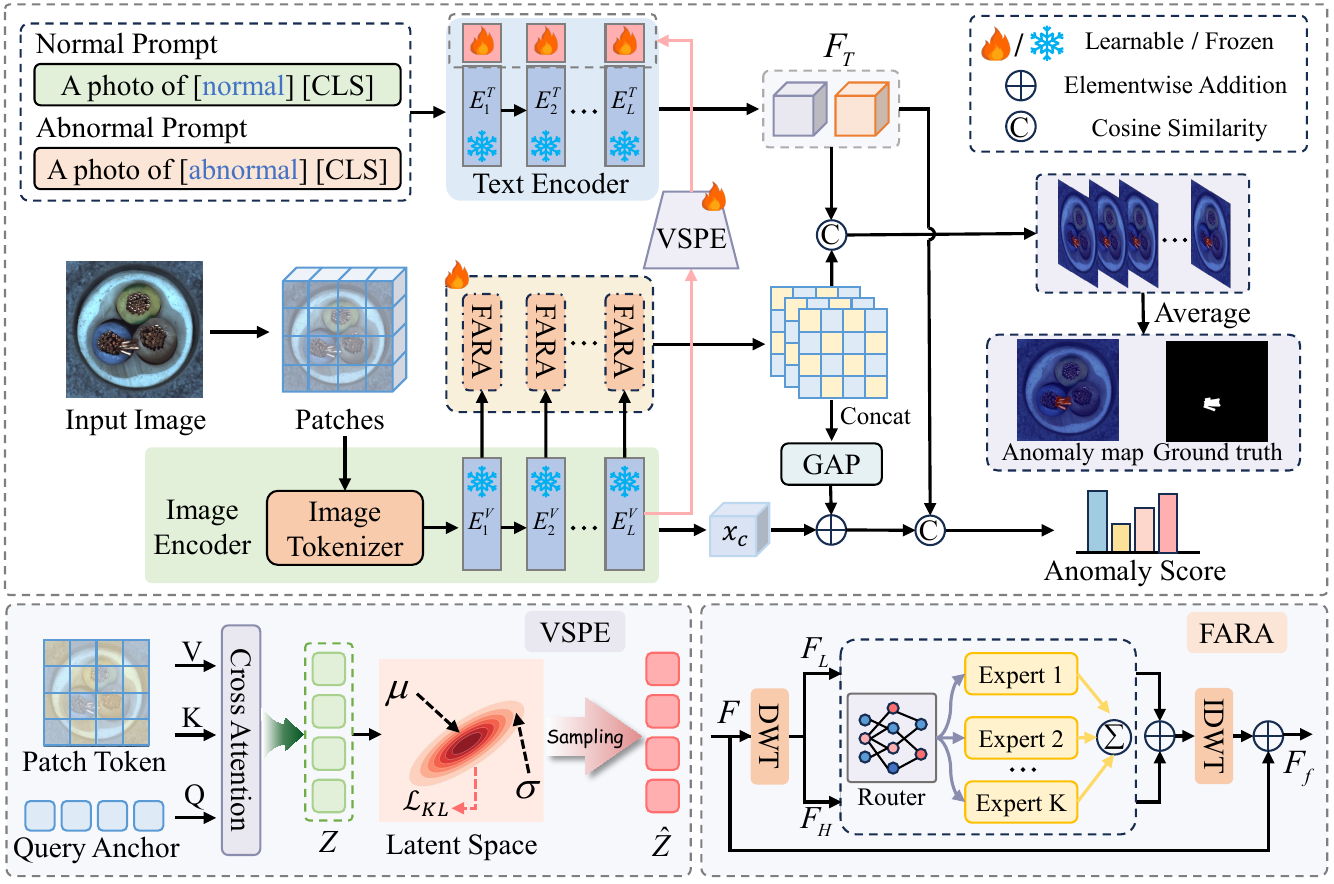}
    \caption{Overview of the proposed VFAD framework. VFAD integrates variational semantic prompting and frequency-adaptive representation learning, where VSPE extracts anomaly-relevant local semantics from dense patch tokens and FARA enhances anomaly-discriminative visual representations via wavelet-based frequency decomposition and expert aggregation.}
    \label{fig:method}
\end{figure*}

\subsection{Problem Definition}
Given an input image, ZSAD aims to identify anomalous patterns in target categories without accessing any category-specific samples during training. Formally, let $\mathcal{C}_{train}$ and $\mathcal{C}_{test}$ denote the category sets used for training and testing, respectively, where $\mathcal{C}_{train} \cap \mathcal{C}_{test} = \varnothing$. For each test image, the model is required to predict an image-level anomaly score $s \in [0,1]$ for anomaly classification, alongside a pixel-level anomaly map $M \in \mathbb{R}^{h \times w}$ for precise anomaly localization. This paradigm inherently demands exceptional zero-shot generalization capabilities, as the model must transfer learned representations to entirely unseen object domains.

\subsection{Overview}
Figure~\ref{fig:method} illustrates the overall framework of VFAD. Given an input image $I \in \mathbb{R}^{h \times w \times 3}$, the image encoder $E^V$ extracts fine-grained patch features. VSPE employs a set of learnable query anchors to aggregate anomaly-relevant local semantics through cross-attention. The resulting representations are regularized by a variational information bottleneck to suppress task-irrelevant redundancy and form discriminative semantic prompts. These prompts are subsequently injected into multiple Transformer layers of the text encoder $E^T$ to facilitate fine-grained cross-modal alignment. In parallel, FARA decomposes the patch features into low- and high-frequency components via a discrete wavelet transform. Frequency-specific MoE branches then adaptively capture complementary global structural and local textural cues, which are aggregated into frequency-aware visual representations. By jointly exploiting variational semantic prompting and frequency-adaptive representation learning, VFAD enhances fine-grained anomaly perception while preserving the zero-shot generalization capability of the pretrained model.

\subsection{Variational Semantic Prompt Extractor}

Although vision-language models exhibit strong zero-shot generalization, their visual representations are primarily optimized for global image-text alignment and are therefore less sensitive to subtle local variations. Directly using dense patch tokens as visual prompts may further introduce redundant background context and noisy local responses, interfering with fine-grained cross-modal alignment. Unlike VCP-CLIP~\cite{qu2024vcp}, which incorporates the global CLS representation to condition text prompts, we propose a Variational Semantic Prompt Extractor (VSPE) that selectively extracts anomaly-relevant local semantics from patch-level features and injects them into the text encoder. This design provides more precise semantic guidance and enhances anomaly-aware visual-textual associations.

Given the patch tokens extracted by the image encoder, denoted as $F \in \mathbb{R}^{N \times C}$, where $N$ and $C$ denote the number of patches and feature dimension, respectively, VSPE introduces learnable query anchors $Q \in \mathbb{R}^{A \times C}$ to aggregate local semantic cues, with $A \ll N$. Specifically, the query anchors interact with dense patch tokens via cross-attention:
\begin{equation}
Z = \mathrm{Attn}(QW_q, FW_k, FW_v),
\end{equation}
where $W_q$, $W_k$, and $W_v$ are learnable projection matrices. Through query-anchor attention, the resulting representation $Z \in \mathbb{R}^{A \times C}$ adaptively aggregates informative local semantics while reducing irrelevant patch responses.

To further reduce redundant information and regularize the extracted prompts, we introduce a variational information bottleneck. Specifically, $Z$ is projected into the parameters of a latent Gaussian distribution:
\begin{equation}
\mu = f_{\mu}(Z), \quad \log \sigma^2 = f_{\sigma}(Z),
\end{equation}
where $f_{\mu}(\cdot)$ and $f_{\sigma}(\cdot)$ are lightweight projection layers. The latent semantic prompts are then sampled using the reparameterization trick:
\begin{equation}
\hat{Z} = \mu + \sigma \odot \epsilon, \quad \epsilon \sim \mathcal{N}(0, I),
\end{equation}
where $\hat{Z} \in \mathbb{R}^{A \times C}$ denotes the variational semantic prompts. During training, the latent distribution is regularized by a Kullback-Leibler divergence term:
\begin{equation}
\mathcal{L}_{\mathrm{KL}} =
-\frac{1}{2}\sum_{i=1}^{A}\sum_{j=1}^{C}
\left(
1 + \log \sigma_{ij}^{2} - \mu_{ij}^{2} - \sigma_{ij}^{2}
\right).
\end{equation}
This regularization constrains the latent prompt distribution and reduces the tendency to overfit category-specific local patterns. The sampled prompts $\hat{Z}$ are further mapped into the text embedding space through a projection layer $\phi(\cdot)$ and combined with learnable prompt embeddings $V_i \in \mathbb{R}^{A \times D}$ to generate text prompt tokens:
\begin{equation}
\mathcal{H}_i= \phi(\hat{Z})+V_i, \quad \mathcal{H}_i\in \mathbb{R}^{A \times D},
\end{equation}
where $D$ is the text embedding dimension. The generated prompt tokens $\mathcal{H}_i$ are then injected into the text Transformer layers by concatenating them with the vanilla text tokens $\mathcal{T}_i \in \mathbb{R}^{P \times D}$, where $P$ denotes the text sequence length. Let $E_i^T$ denote the $i$-th Transformer layer of the text encoder. The prompt-enhanced text encoding process is formulated as:
\begin{align}
[\mathcal{T}_{i+1}, \_] &= E_i^T([\mathcal{T}_i, \mathcal{H}_i]), \quad i \le J, \\
[\mathcal{T}_{i+1}, \mathcal{H}_{i+1}] &= E_i^T([\mathcal{T}_i, \mathcal{H}_i]), \quad i > J,
\end{align}
where $J$ denotes the maximum depth for prompt updating, and $[\cdot,\cdot]$ represents concatenation along the sequence dimension. By injecting variational semantic prompts into the text encoder, VSPE enables textual representations to incorporate anomaly-relevant local visual semantics, thereby enhancing fine-grained visual-textual alignment for ZSAD.

\subsection{Frequency-Adaptive Representation Aggregation}

Anomalies often manifest as subtle texture variations or localized structural distortions, which are difficult to capture sufficiently using spatial-domain patch representations alone. Although the CLIP image encoder provides strong semantic representations, its global feature aggregation tends to emphasize object-level semantics while attenuating fine-grained local variations, thereby limiting its sensitivity to high-frequency anomaly cues. To address this limitation, we propose Frequency-Adaptive Representation Aggregation (FARA), which explicitly decomposes visual features into complementary frequency components and adaptively refines them using frequency-specific experts.

We first reshape the patch features $F \in \mathbb{R}^{N \times C}$ into a spatial feature map and apply a discrete wavelet transform (DWT) to decompose them into one low-frequency subband $F_{LL}$ and three directional high-frequency subbands $F_{LH}$, $F_{HL}$, and $F_{HH}$. The resulting low- and high-frequency representations are formulated as:
\begin{align}
F_L=F_{LL}, \quad F_H=[F_{LH}, F_{HL}, F_{HH}],
\end{align}
where $F_L$ encodes global structural information, while $F_H$ contains local details. This decomposition enables the model to separately characterize structural and textural variations that are entangled in the original spatial representation.

To adaptively enhance frequency-specific representations, we introduce two independent MoE branches for the low- and high-frequency components, respectively. Each branch contains $K$ frequency-specific experts, while a lightweight router dynamically activates the top-$k$ experts according to the corresponding input representation. For each frequency component $F_r$, where $r\in\{L,H\}$, the routing scores and activated expert indices are computed as:
\begin{equation}
s_r=\mathcal{R}_r(F_r)\in\mathbb{R}^{K},
\qquad
\mathcal{I}_r=\operatorname{TopK}(s_r,k),
\end{equation}
where $\mathcal{R}_r(\cdot)$ denotes the router of branch $r$, and $\mathcal{I}_r$ denotes the index set of the selected top-$k$ experts. The routing weights are then normalized over the activated experts:
\begin{equation}
\alpha_{r,j}
=
\begin{cases}
\displaystyle
\frac{\exp(s_{r,j})}
{\sum_{m\in\mathcal{I}_r}\exp(s_{r,m})},
& j\in\mathcal{I}_r, \\[8pt]
0,
& j\notin\mathcal{I}_r.
\end{cases}
\end{equation}

The enhanced frequency representation is obtained by sparsely aggregating the outputs of the selected experts:
\begin{equation}
\hat{F}_r
=
\sum_{j\in\mathcal{I}_r}
\alpha_{r,j}\,
\mathcal{E}_{r,j}(F_r),
\end{equation}
where $\mathcal{E}_{r,j}(\cdot)$ denotes the $j$-th expert in branch $r$. This sparse routing mechanism enables the model to adaptively select specialized experts for diverse structural and textural patterns.
The enhanced frequency components are then transformed back into the spatial domain through an inverse discrete wavelet transform. To preserve the original representation and mitigate over-adaptation to auxiliary training categories, we adopt a residual connection:
\begin{equation}
F_f = F + \mathrm{IDWT}(\hat{F}_L, \hat{F}_H).
\end{equation}






\begin{table*}[]
\centering
\fontsize{9}{11}\selectfont
\setlength{\tabcolsep}{1.55mm}
\begin{tabular}{ccccccccc}
\toprule
Metric & Dataset 
& \begin{tabular}[c]{@{}c@{}}WinCLIP\\CVPR 2023\end{tabular}
& \begin{tabular}[c]{@{}c@{}}AnomalyCLIP\\ICLR 2024\end{tabular}
& \begin{tabular}[c]{@{}c@{}}AdaCLIP\\ECCV 2024\end{tabular}
& \begin{tabular}[c]{@{}c@{}}AA-CLIP\\CVPR 2025\end{tabular}
& \begin{tabular}[c]{@{}c@{}}Bayes-PFL\\CVPR 2025\end{tabular}
& \begin{tabular}[c]{@{}c@{}}MoECLIP\\CVPR 2026\end{tabular}
& \begin{tabular}[c]{@{}c@{}}VFAD\\Ours\end{tabular} \\ \cline{3-9}
\hline
\multirow{7}{*}{\begin{tabular}[c]{@{}c@{}}Image-level\\ (AUROC, AP)\end{tabular}} 
& VisA & (78.1, 77.5) & (82.1, 85.4) & (83.0, 84.9) & (84.6, 83.4) & (\underline{86.8}, \underline{89.3}) & (83.6, 86.2) & (\textbf{87.2}, \textbf{89.7}) \\
& MVTec-AD & (91.8, 95.1) & (91.5, 96.2) & (92.0, 96.4) & (90.5, 95.6) & (92.2, 96.1) & (\textbf{93.9}, \underline{96.8}) & (\underline{93.1}, \textbf{96.9}) \\
& KSDD2 & (93.5, 77.9) & (92.1, 77.8) & (95.9, 95.9) & (95.8, 96.8) & (\underline{97.3}, \underline{97.9}) & (94.6, 94.1) & (\textbf{97.5}, \textbf{98.1}) \\
& DAGM & (89.6, 90.4) & (97.5, 92.3) & (96.5, 92.7) & (94.3, 84.2) & (\underline{97.7}, \underline{97.0}) & (95.2, 87.5) & (\textbf{97.9}, \textbf{97.1}) \\
& DTD-Synthetic & (93.2, 92.6) & (93.5, 97.0) & (92.8, 97.0) & (91.4, 97.2) & (93.5, 97.7) & (\textbf{95.5}, \underline{98.6}) & (\underline{95.3}, \textbf{98.8}) \\ \cline{2-9} 
& Average & (89.2, 86.7) & (91.3, 89.7) & (92.0, 93.4) & (91.3, 91.4) & (\underline{93.5}, \underline{95.6}) & (92.6, 92.6) & (\textbf{94.2}, \textbf{96.1}) \\ \hline

\multirow{7}{*}{\begin{tabular}[c]{@{}c@{}}Pixel-level\\ (AUROC, AP)\end{tabular}} 
& VisA & (79.6, \phantom{1}5.0) & (95.5, 21.3) & (95.1, \textbf{29.2}) & (95.5, 25.0) & (95.5, \textbf{29.2}) & (\underline{95.6}, 26.1) & (\textbf{95.8}, \underline{28.7}) \\
& MVTec-AD & (85.1, 18.0) & (91.1, 34.5) & (86.8, 38.1) & (91.9, 45.4) & (91.9, \underline{48.4}) & (\underline{92.5}, 45.7) & (\textbf{92.7}, \textbf{48.6}) \\
& KSDD2 & (97.9, 17.1) & (99.4, 41.8) & (96.1, 56.4) & (99.5, 43.7) & (\underline{99.6}, \underline{73.7}) & (98.1, 53.8) & (\textbf{99.7}, \textbf{73.9}) \\
& DAGM & (83.2, \phantom{1}3.1) & (95.6, 29.5) & (91.5, 44.2) & (96.3, \underline{45.9}) & (\underline{99.3}, 43.1) & (94.7, 43.9) & (\textbf{99.5}, \textbf{46.2}) \\
& DTD-Synthetic & (83.9, 11.6) & (97.9, 52.4) & (94.1, 52.8) & (96.4, 62.8) & (97.8, \underline{69.9}) & (\underline{98.8}, 62.7) & (\textbf{98.9}, \textbf{71.4}) \\ \cline{2-9} 
& Average & (85.9, 11.0) & (95.9, 35.9) & (92.7, 44.1) & (95.9, 44.6) & (\underline{96.8}, \underline{52.9}) & (95.9, 46.4) & (\textbf{97.3}, \textbf{53.8}) \\ \bottomrule
\end{tabular}
\caption{ZSAD performance comparison with state-of-the-art methods on the industrial domain. The best results are marked in bold, and the second-best are underlined.}
\label{table_industrial}
\end{table*}

\subsection{Anomaly Map and Anomaly Score}
\subsubsection{Pixel-level anomaly map.}
Given an input image $I \in \mathbb{R}^{h \times w \times 3}$, the image encoder extracts patch-level visual features from $L$ intermediate layers, where the feature from the $l$-th layer is denoted as $F^l \in \mathbb{R}^{N \times C}$. Each layer feature is further refined by the proposed FARA module to obtain the enhanced visual representation $F_f^l \in \mathbb{R}^{N \times C}$. Meanwhile, the text encoder generates the normal and anomalous textual embeddings, denoted as $F_T=[F_T^n,F_T^a]\in \mathbb{R}^{2 \times C}$. The final pixel-level anomaly map $M\in \mathbb{R}^{h \times w}$ is obtained by averaging the anomalous maps from all selected layers:
\begin{equation}
M = \frac{1}{L}\sum_{l=1}^{L}
\left[
\mathrm{Softmax}\left(
\mathrm{Up}\left(\tilde{F}_f^l \tilde{F}_T^{\top}\right)
\right)
\right],
\end{equation}
where $\tilde{(\cdot)}$ denotes $L_2$ normalization along the embedding dimension, $\mathrm{Up}(\cdot)$ denotes the upsampling operation.

\subsubsection{Image-level anomaly score.}

For image-level anomaly classification, we aggregate multi-level visual cues into an image-level representation and align it with textual semantics. Specifically, we first concatenate $F_f^l \in \mathbb{R}^{N \times C}$ along the channel dimension to integrate multi-level representations. The resulting feature is then aggregated over spatial dimensions via global average pooling, followed by a linear projection that maps it back to the original feature dimension. To preserve the global semantic prior, we further fuse the pooled patch representation with the class token $x_c$ extracted from the original image encoder:
\begin{equation}
G_I = x_c + \mathrm{Linear}\left(\mathrm{GAP}\left(\delta[F_f^1,F_f^2,\ldots,F_f^L]\right)\right),
\end{equation}
where $G_I \in \mathbb{R}^{C}$ denotes the resulting image-level visual representation, $\delta[\cdot]$ denotes channel-wise concatenation, and $\mathrm{GAP}(\cdot)$ performs global average pooling over spatial dimensions. The image-level anomaly score is computed by measuring the similarity between the image-level representation and textual embeddings:
\begin{equation}
S_I =
\mathrm{Softmax}\left(
\tilde{G}_I\tilde{F}_T^{\top}
\right).
\end{equation}

\subsection{Loss Function}
During training, we jointly optimize the model using segmentation and classification losses. The segmentation loss combines Dice loss and Focal loss to supervise pixel-level anomaly localization:
\begin{equation}
\mathcal{L}_{seg} = \sum_{l=1}^L \Big( \mathrm{Dice}(M^l, M_{gt}) + \mathrm{Focal}(M^l, M_{gt}) \Big),
\end{equation}
where $M^l$ denotes the predicted anomaly map from the $l$-th layer, and $M_{gt}$ is the corresponding ground-truth mask.
For image-level anomaly classification, we adopt binary cross-entropy loss to supervise the predicted anomaly score $S_I$ using the image-level ground-truth label $S_{gt}$. The overall training objective is formulated as:
\begin{equation}
\mathcal{L}_{total} =
\mathrm{BCE}(S_I, S_{gt})
+ \mathcal{L}_{seg}
+ \lambda \mathcal{L}_{KL},
\end{equation}
where $\lambda$ is a hyperparameter that balances the contribution of the KL divergence regularization term.

\begin{table*}[]
\centering
\fontsize{9}{11}\selectfont
\setlength{\tabcolsep}{1.7mm}
\begin{tabular}{ccccccccc}
\toprule
Metric & Dataset 
& \begin{tabular}[c]{@{}c@{}}WinCLIP\\CVPR 2023\end{tabular}
& \begin{tabular}[c]{@{}c@{}}AnomalyCLIP\\ICLR 2024\end{tabular}
& \begin{tabular}[c]{@{}c@{}}AdaCLIP\\ECCV 2024\end{tabular}
& \begin{tabular}[c]{@{}c@{}}AA-CLIP\\CVPR 2025\end{tabular}
& \begin{tabular}[c]{@{}c@{}}Bayes-PFL\\CVPR 2025\end{tabular}
& \begin{tabular}[c]{@{}c@{}}MoECLIP\\CVPR 2026\end{tabular}
& \begin{tabular}[c]{@{}c@{}}VFAD\\Ours\end{tabular} \\ \cline{3-9}
\hline
\multirow{4}{*}{\begin{tabular}[c]{@{}c@{}}Image-level\\ (AUROC, AP)\end{tabular}} 
& HeadCT   & (83.7, 81.6) & (95.3, 95.2) & (93.3, 92.2) & (\underline{96.8}, 95.3) & (96.5, \underline{95.5}) & (96.6, 94.5) & (\textbf{98.3}, \textbf{98.0}) \\
& BrainMRI & (92.0, 90.7) & (96.1, 92.3) & (94.9, 94.2) & (94.1, 93.3) & (\underline{96.2}, 92.4) & (88.5, \underline{97.1}) & (\textbf{96.3}, \textbf{97.2}) \\
& Br35H    & (80.5, 82.2) & (97.3, 96.1) & (95.7, 95.7) & (96.2, 94.2) & (\underline{97.8}, \underline{96.2}) & (96.0, 93.8) & (\textbf{98.3}, \textbf{97.7}) \\ \cline{2-9}
& Average  & (85.4, 84.8) & (96.2, 94.5) & (94.6, 94.0) & (95.7, 94.3) & (\underline{96.8}, 94.7) & (93.7, \underline{95.1}) & (\textbf{97.6}, \textbf{97.6}) \\ \hline

\multirow{6}{*}{\begin{tabular}[c]{@{}c@{}}Pixel-level\\ (AUROC, AP)\end{tabular}} 
& ISIC     & (83.3, 62.4) & (88.4, 74.4) & (85.4, 70.6) & (\underline{93.8}, \underline{85.9}) & (92.2, 84.6) & (91.7, 83.9) & (\textbf{94.2}, \textbf{86.4}) \\
& ColonDB  & (64.8, 14.3) & (81.9, 31.3) & (79.3, 26.2) & (83.6, 32.1) & (82.1, 31.9) & (\textbf{85.4}, \underline{34.8}) & (\underline{84.1}, \textbf{36.0}) \\
& ClinicDB & (70.7, 19.4) & (85.9, 42.2) & (84.3, 36.0) & (89.5, \underline{56.5}) & (89.6, 53.2) & (\underline{89.7}, 49.9) & (\textbf{90.1}, \textbf{57.2}) \\
& Endo     & (68.2, 23.8) & (86.3, 50.4) & (84.0, 44.8) & (90.2, 60.7) & (89.2, 58.6) & (\underline{91.0}, \underline{62.5}) & (\textbf{91.7}, \textbf{64.1}) \\
& Kvasir   & (69.8, 27.5) & (81.8, 42.5) & (79.4, 43.8) & (87.6, 57.2) & (85.4, 54.2) & (\underline{88.1}, \underline{57.6}) & (\textbf{88.2}, \textbf{60.0}) \\ \cline{2-9}
& Average  & (71.4, 29.5) & (84.9, 48.2) & (82.5, 44.3) & (89.0, \underline{58.5}) & (87.7, 56.5) & (\underline{89.2}, 57.7) & (\textbf{89.7}, \textbf{60.7}) \\ \bottomrule
\end{tabular}
\caption{ZSAD performance comparison with state-of-the-art methods on the medical domain. The best results are marked in bold, and the second-best are underlined.}
\label{table_medical}
\end{table*}

\begin{figure}
    \centering
    \includegraphics[width=\linewidth]{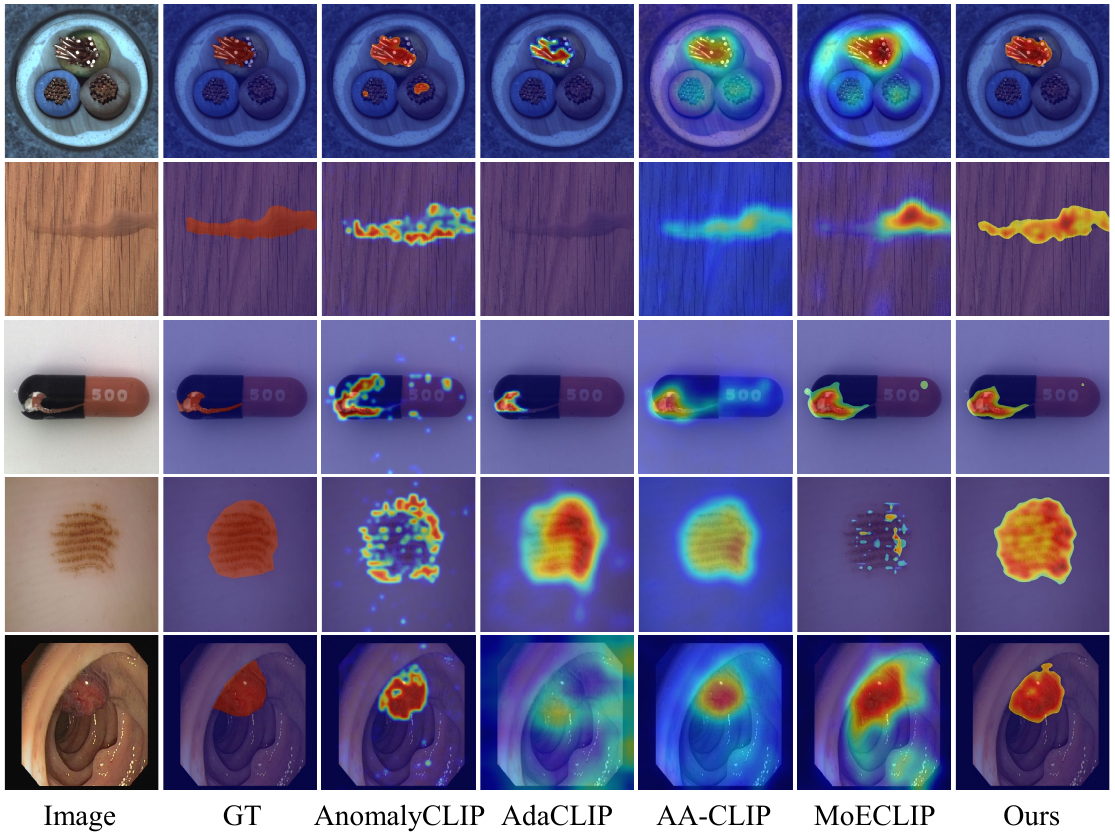}
    \caption{Visualization of anomaly localization results of AnomalyCLIP, AdaCLIP, AA-CLIP, MoECLIP and ours.}
    \label{fig:vis}
\end{figure}

\section{Experiments}
\subsection{Experimental Setups}
\subsubsection{Datasets.} 
To provide a comprehensive evaluation of the ZSAD performance of our model, we conduct experiments on 13 public anomaly detection benchmarks across both industrial and medical domains. In the industrial domain, we adopt five representative benchmarks, including MVTec-AD~\cite{bergmann2019mvtec}, VisA~\cite{zou2022spot}, KSDD2~\cite{bovzivc2021mixed}, DAGM~\cite{wieler2007weakly}, and DTD-Synthetic~\cite{aota2023zero}. In the medical domain, we evaluate our model on eight datasets, including HeadCT~\cite{salehi2021multiresolution}, BrainMRI~\cite{kanade2015brain}, and Br35H~\cite{hamada2020br35h} for brain tumor classification, ISIC~\cite{codella2018skin} for skin lesion detection, and CVC-ColonDB~\cite{tajbakhsh2015automated}, CVC-ClinicDB~\cite{bernal2015wm}, Endo~\cite{hicks2021endotect}, and Kvasir~\cite{jha2019kvasir} for colon polyp segmentation.

\subsubsection{Evaluation metrics.}
We evaluate the ZSAD performance from both image-level anomaly classification and pixel-level anomaly segmentation perspectives. Following existing works~\cite{park2026moeclip}, we adopt the Area Under the Receiver Operating Characteristic curve (AUROC) and Average Precision (AP) as the primary evaluation metrics. 

\subsubsection{Implementation details.}
We adopt the publicly available CLIP model with ViT-L/14-336 as the default backbone. All input images are resized to $518 \times 518$. Patch features are extracted from the 6th, 12th, 18th, and 24th layers of the visual encoder. The prompting depth $J$ and the number of learnable query anchors $A$ are set to 9 and 8, respectively. Each MoE branch comprises four experts and employs top-2 routing. The KL regularization weight $\lambda$ is set to 0.1. We optimize the model using Adam with a learning rate of $2\times10^{-4}$. Following prior works~\cite{qu2025bayesian,park2026moeclip}, we use VisA as the auxiliary training set since its categories do not overlap with those of the remaining datasets. For evaluation on VisA, we train the model on MVTec-AD. Experiments are conducted on a single NVIDIA A100 GPU. More details are provided in the Supplementary Material A.

\begin{table}[t]
\centering
\fontsize{9}{11}\selectfont
\setlength{\tabcolsep}{1.2mm}
\begin{tabular}{cc|cc|cc}
\toprule
\multirow{2}{*}{VSPE} &
\multirow{2}{*}{FARA} &
\multicolumn{2}{c|}{MVTec-AD} &
\multicolumn{2}{c}{VisA} \\
\cmidrule(lr){3-4}
\cmidrule(l){5-6}
& &
I-AUROC &
P-AUROC &
I-AUROC &
P-AUROC \\
\midrule
\ding{55} & \ding{55} & 89.3 & 90.6 & 84.4 & 94.7 \\
\ding{51} & \ding{55} & 91.5 & 91.2 & 86.3 & 95.4 \\
\ding{55} & \ding{51} & 90.9 & 91.8 & 85.5 & 95.2 \\
\ding{51} & \ding{51} &
\textbf{93.1} &
\textbf{92.7} &
\textbf{87.2} &
\textbf{95.8} \\
\bottomrule
\end{tabular}
\caption{Ablation study of the proposed components on MVTec-AD and VisA. I-AUROC and P-AUROC represent Image-level and Pixel-level AUROC, respectively.}
\label{tab:ablation}
\end{table}

\subsection{Comparison with State-of-the-Art Methods}
We compare VFAD against six representative state-of-the-art ZSAD methods: WinCLIP~\cite{jeong2023winclip}, AnomalyCLIP~\cite{zhou2024anomalyclip}, AdaCLIP~\cite{cao2024adaclip}, AA-CLIP~\cite{ma2025aa}, Bayes-PFL~\cite{qu2025bayesian}, and MoECLIP~\cite{park2026moeclip}. For a fair comparison, all methods are evaluated under a unified experimental protocol. More results are provided in the Supplementary Material B.

\begin{figure}
    \centering
    \includegraphics[width=\linewidth]{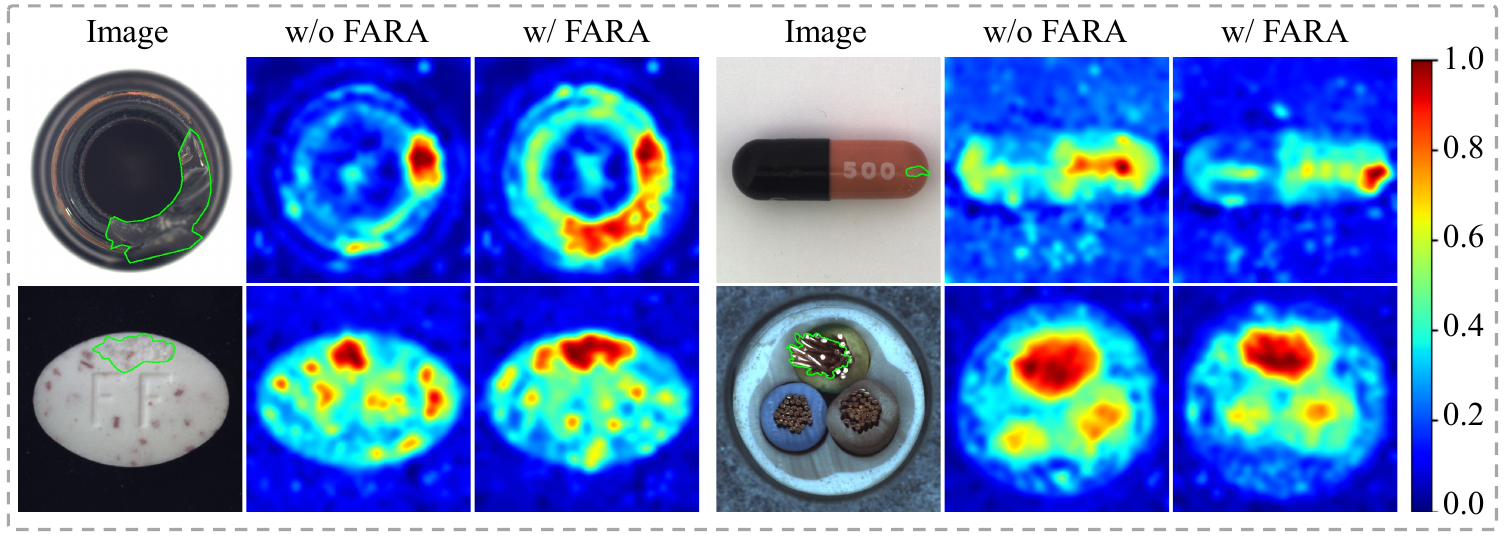}
    \caption{Qualitative comparison of anomaly localization results without and with the FARA module.}
    \label{fig:fara}
\end{figure}

\subsubsection{Quantitative comparison.}
As shown in Table~\ref{table_industrial}, VFAD achieves the best average performance on industrial benchmarks, reaching 94.2\% AUROC and 96.1\% AP for image-level classification, and 97.3\% AUROC and 53.8\% AP for pixel-level localization. Compared with the strongest competitor, it improves the average image-level and pixel-level AUROC by 0.7\% and 0.5\%, respectively, while ranking first on nearly all datasets, demonstrating robust performance across diverse object categories and anomaly patterns. Table~\ref{table_medical} further presents the comparison on medical benchmarks. VFAD achieves state-of-the-art performance, with average image-level scores of 97.6\% AUROC and 97.6\% AP, together with average pixel-level scores of 89.7\% AUROC and 60.7\% AP. In particular, the improvements in AP indicate more reliable anomaly discrimination and more accurate lesion localization. These consistent gains demonstrate the strong generalization capability of the proposed framework.

\subsubsection{Qualitative comparisons.} 
As shown in Figure~\ref{fig:vis}, VFAD consistently produces more accurate anomaly maps across both industrial and medical datasets. Compared with existing methods, VFAD more effectively suppresses spurious background activations while retaining subtle anomaly cues, producing sharper boundaries and more complete anomalous regions. These qualitative results further demonstrate the effectiveness of VFAD in capturing fine-grained anomaly patterns and its strong generalization across diverse domains.

\subsection{Ablation Study}

\subsubsection{Effect of individual components.}

Table~\ref{tab:ablation} presents the ablation results of VSPE and FARA on MVTec-AD and VisA. Both modules consistently improve the baseline, with VSPE mainly enhancing image-level anomaly discrimination through anomaly-relevant semantic modeling, while FARA contributes more to pixel-level localization by strengthening frequency-aware visual representations. Their combination achieves the best performance across all metrics, reaching 93.1\% I-AUROC and 92.7\% P-AUROC on MVTec-AD, and 87.2\% I-AUROC and 95.8\% P-AUROC on VisA. As further illustrated in Figure~\ref{fig:fara}, FARA produces more complete and concentrated anomaly responses, with clearer boundaries and reduced background interference. These results highlight the complementary roles of VSPE and FARA in anomaly discrimination and fine-grained localization.

\subsubsection{Influence of VSPE.}
As shown in Table~\ref{tab:vspe_ablation}, we investigate the effects of the semantic source and variational information bottleneck in VSPE. Local semantic aggregation consistently outperforms the use of the global token, demonstrating that dense patch tokens provide more informative cues for fine-grained anomaly modeling. Incorporating VIB further improves both variants by regularizing the semantic prompt space and suppressing redundant information.

\begin{table}[t]
\centering
\fontsize{9}{11}\selectfont
\setlength{\tabcolsep}{0.7mm}
\begin{tabular}{ccc|cc|cc}
\toprule
\multicolumn{2}{c}{Semantic} &
\multirow{2}{*}{VIB} &
\multicolumn{2}{c|}{MVTec-AD} &
\multicolumn{2}{c}{VisA} \\
\cmidrule(lr){1-2}
\cmidrule(lr){4-5}
\cmidrule(lr){6-7}
Global &
Local &
&
I-AUROC &
P-AUROC &
I-AUROC &
P-AUROC \\
\midrule
\ding{51} & \ding{55} & \ding{55} &
90.2 & 91.1 & 84.7 & 94.9 \\
\ding{55} & \ding{51} & \ding{55} &
91.9 & 91.8 & 86.1 & 95.4 \\
\ding{51} & \ding{55} & \ding{51} &
91.1 & 91.7 & 85.8 & 95.3 \\
\ding{55} & \ding{51} & \ding{51} &
\textbf{93.1} &
\textbf{92.7} &
\textbf{87.2} &
\textbf{95.8} \\
\bottomrule
\end{tabular}
\caption{Ablation study of the semantic source and variational information bottleneck in VSPE on MVTec-AD and VisA.}
\label{tab:vspe_ablation}
\end{table}

\begin{figure}
    \centering
    \includegraphics[width=\linewidth]{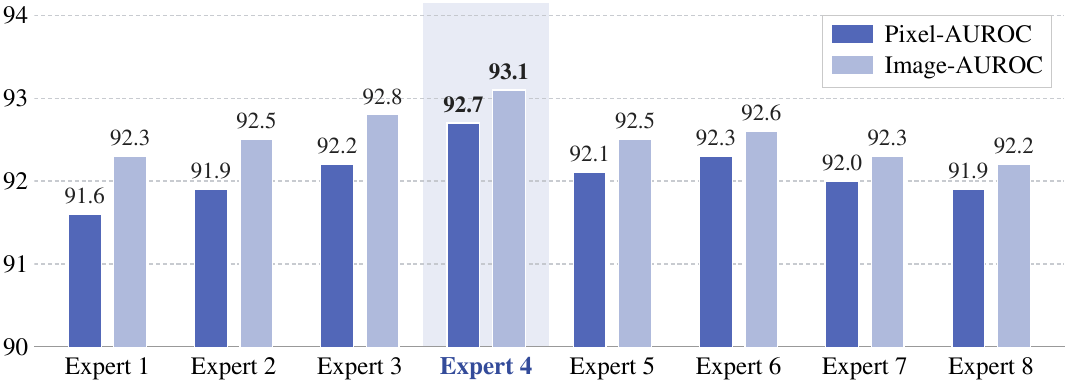}
    \caption{Influence of the number of experts in each frequency-specific MoE branch on the MVTec-AD dataset.}
    \label{fig:expert}
\end{figure}

\subsubsection{Influence of the number of experts.}
We investigate the effect of the number of experts in each frequency-specific MoE branch on MVTec-AD. As shown in Figure~\ref{fig:expert}, increasing the number of experts from 1 to 4 steadily improves both image-level and pixel-level AUROC, indicating that multiple experts facilitate the learning of complementary frequency representations. The best performance is achieved with four experts. Further increasing the number of experts results in slight performance degradation, possibly due to increased expert redundancy and optimization difficulty.

\subsubsection{Influence of different CLIP backbones.}
Table~\ref{tab:backbone} compares different CLIP backbone configurations on MVTec-AD. Increasing the model capacity consistently improves both image-level classification and pixel-level localization performance. ViT-L/14-336 achieves the best overall performance, with 93.1\% I-AUROC and 92.7\% P-AUROC, indicating that a stronger backbone and higher-resolution pre-training better capture discriminative semantics and fine-grained anomaly details. Although ViT-L/14-336 introduces
additional inference latency, it provides the best overall performance and is therefore adopted as the default backbone.

\begin{table}
\centering
\fontsize{9}{11}\selectfont
\setlength{\tabcolsep}{1mm}
\begin{tabular}{cccccc}
\toprule
Backbone &
I-AUROC &
I-AP &
P-AUROC &
P-AP &
Time (ms) \\
\midrule
ViT-B/16-224 & 90.7 & 93.8 & 90.3 & 45.3 & \textbf{91.6} \\
ViT-L/14-224 & 91.9 & 95.2 & 91.8 & 46.7 & 112.9 \\
ViT-L/14-336 & \textbf{93.1} & \textbf{96.9} & \textbf{92.7} & \textbf{48.6} & 137.5 \\
\bottomrule
\end{tabular}
\caption{Comparison of different pretrained backbones.}
\label{tab:backbone}
\end{table}


\begin{figure}
    \centering
    \includegraphics[width=\linewidth]{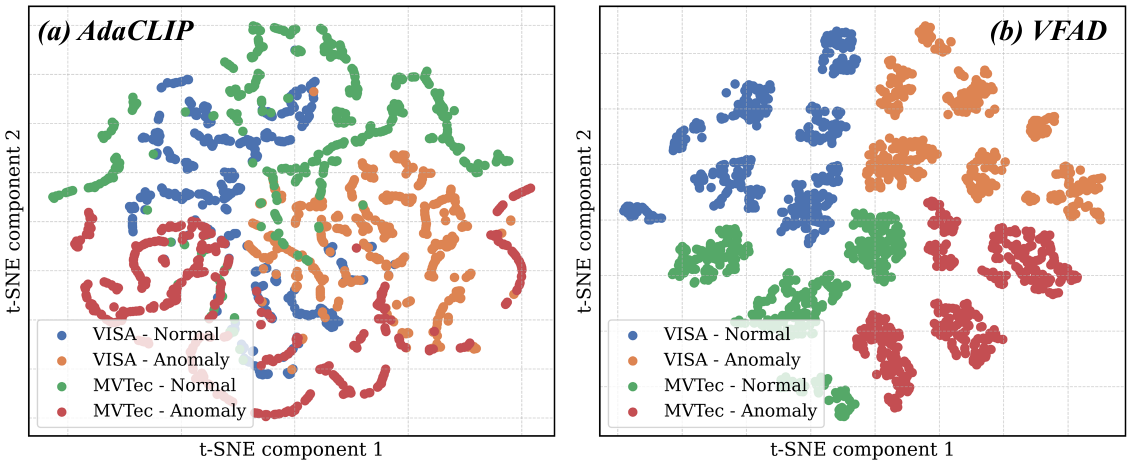}
    \caption{t-SNE visualization of text feature distributions on the MVTec-AD and VisA datasets.}
    \label{fig:tsne}
\end{figure}
\subsubsection{Visualization of the text embedding space.} 
Figure~\ref{fig:tsne} presents the t-SNE visualization of text embeddings generated by AdaCLIP and our method. Compared with AdaCLIP, our method forms more compact semantic clusters and achieves clearer separation between normal and anomalous semantics. The embeddings produced by AdaCLIP remain relatively scattered and partially mixed, whereas our method exhibits improved semantic separability. This observation indicates that the proposed VSPE provides more discriminative anomaly-aware semantic guidance by aggregating informative local cues and suppressing redundant visual information.

\section{Conclusion}

In this work, we presented VFAD, a zero-shot anomaly detection framework that integrates variational semantic prompting with frequency-adaptive representation learning. VSPE extracts anomaly-relevant local semantics from dense patch features and regularizes their latent distribution through a variational information bottleneck, thereby incorporating fine-grained visual cues into cross-modal alignment. FARA further strengthens texture-sensitive visual representations through frequency decomposition and adaptive expert aggregation. Extensive experiments demonstrate the superior performance of VFAD. Future work will explore more efficient prompt modeling and broader real-world applications.

\bibliography{aaai2027}

\end{document}